\documentclass{article}

\usepackage[nonatbib, final]{nips_2017}

\usepackage[utf8]{inputenc} 
\usepackage[T1]{fontenc}    
\usepackage{hyperref}       
\usepackage{url}            
\usepackage{booktabs}       
\usepackage{amsfonts}       
\usepackage{nicefrac}       
\usepackage{microtype}      

\usepackage[sort&compress,square,numbers]{natbib}
\usepackage{amsmath,amssymb,amsfonts,amsthm}
\usepackage{graphicx}
\usepackage{adjustbox}
\usepackage{multirow}
\usepackage{wrapfig}
\usepackage{color, colortbl}
\usepackage{arydshln}
\usepackage[font=small,labelfont=bf]{caption}
\usepackage{subcaption}
\definecolor{Gray}{gray}{0.9}
\usepackage{etoolbox}
\newcommand{\zerodisplayskips}{%
  \setlength{\abovedisplayskip}{2pt}%
  \setlength{\belowdisplayskip}{2pt}%
  \setlength{\abovedisplayshortskip}{0pt}%
  \setlength{\belowdisplayshortskip}{0pt}}
\appto{\normalsize}{\zerodisplayskips}
\appto{\small}{\zerodisplayskips}
\appto{\footnotesize}{\zerodisplayskips}

\title{Recurrent Inference Machines \\
for Solving Inverse Problems}

\author{Patrick Putzky \& Max Welling \\
Informatics Institute\\
University of Amsterdam\\
\texttt{\{pputzky,m.welling\}@uva.nl}
}

\newcommand{\figref}[1]{\figurename~\ref{#1}}

\newcommand{\x}{\mathbf{x}}
\newcommand{\y}{\mathbf{y}}

\newcommand{\A}{\mathbf{A}}

\newcommand{\z}{\boldsymbol{\eta}}
\newcommand{\n}{\mathbf{n}}
\newcommand{\s}{\mathbf{s}}

\begin{document}

\maketitle

\begin{abstract}
Much of the recent research on solving iterative inference problems focuses on moving away from hand-chosen inference algorithms and towards learned inference. In the latter, the inference process is unrolled in time and interpreted as a recurrent neural network (RNN) which allows for joint learning of model and inference parameters with back-propagation through time. In this framework, the RNN architecture is directly derived from a hand-chosen inference algorithm, effectively limiting its capabilities.
We propose a learning framework, called \textit{Recurrent Inference Machines (RIM)}, in which we turn algorithm construction the other way round: Given data and a task, train an RNN to learn an inference algorithm. Because RNNs are Turing complete \cite{Siegelmann1991,Siegelmann1995} they are capable to implement any inference algorithm. The framework allows for an abstraction which removes the need for domain knowledge.
We demonstrate in several image restoration experiments that this abstraction is effective, allowing us to achieve state-of-the-art performance on image denoising and super-resolution tasks and superior across-task generalization.
\end{abstract}

\section{Introduction}\label{sec:Intro}
Inverse Problems are a broad class of problems which can be encountered in all scientific disciplines, from the natural sciences to engineering. The task in inverse problems is to reconstruct a signal from observations that are subject to a known (or inferred) corruption process known as the forward model. In this work we will focus on linear measurement problems of the form
\begin{align}\label{eq:linear_mesaurement}
\y = \A \x + \n ,
\end{align}
where $\y$ is a noisy measurement vector, $\x$ is the signal of interest, $\A$ is an $m \times d$ corruption matrix, and $\n$ is an additive noise vector. If $\A$ is a wide matrix such that $m \ll d$, this problem is typically ill-posed. Many signal reconstruction problems can be phrased in terms of the linear measurement problem such as image denoising, super-resolution, and deconvolution. The general form of $\A$ typically defines the problem class. If $\A$ is an identity matrix the problem is a denoising problem, while in tomography $\A$ represents a Fourier transform and a consecutive sub-sampling of the Fourier coefficients. In this paper we assume the forward model is known.

One way to approach inverse problems is by defining a likelihood and prior, and optimizing for the maximum a posteriori (MAP) solution \citep{Figueiredo2007}:
\begin{align} \label{eq:map_optimization}
\max_{\x} \log p(\y|\x)  + \log p_\theta(\x)
\end{align}
Here, $p(\y|\x)$ is a likelihood term representing the noisy forward model, and $p_\theta(\x)$ is a parametric prior over $\x$ which reduces the solution space for an otherwise ill-posed problem. In classical optimization frameworks there is a trade-off between expressiveness of the prior $p_\theta(\x)$ and optimization performance. While more expressive priors allow for better representation of the signal of interest, they will typically make optimization more difficult. In fact, only for a few trivial prior-likelihood pairs will inference remain convex. In practice one often has to resort to approximations of the objective and to approximate double-loop algorithms in order to allow for scalable inference \citep{Nickisch2009,Zoran2011}.

In this work we take a different approach to solving inverse problems. We move away from the idea that it is beneficial to separate learning a prior (or regularizer) from the optimization procedure to do the reconstruction. The usual thinking is that this separation allows for greater modularity and the possibility to interchange one of these two complementary components in order to build new algorithms. In practice however, we observe that the optimization procedure almost always has to be adapted to the model choice to achieve good performance   \citep{Aharon2006,Elad2006,Nickisch2009,Zoran2011}. In fact, it is well known that the optimization procedure used for training should match the one used during testing because the model has adapted itself to perform well under that optimization procedure \citep{Kumar2005, Wainwright2006}.
\begin{wrapfigure}[14]{r}{0.4\textwidth}
  \begin{center}
  \vspace{-10pt}
    \includegraphics[width=0.4\textwidth]{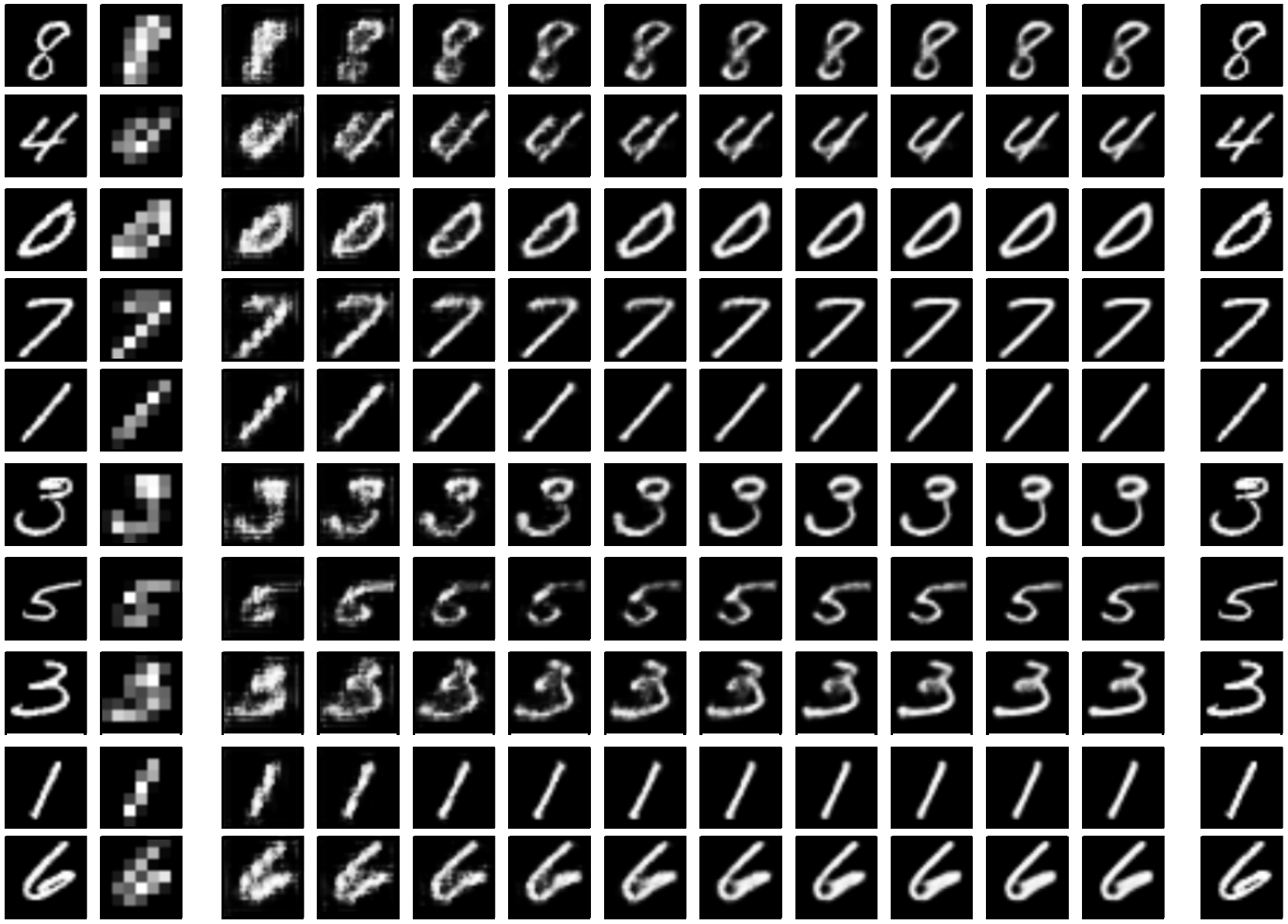}
  \end{center}
  \caption{\textit{An RIM performs factor-4 super-resolution task on MNIST.}}
\end{wrapfigure}
In line with the idea of jointly learning prior and inference, research in recent years has moved towards interpreting iterative inference algorithms as recurrent neural networks with a set of inference and model parameters that can be learned jointly with back-propagation through time \cite{Chen2015,Gregor2010,Wang2016,Zheng2015}. The approach allows model and inference parameters to flexibly adapt to the data and the problem task. This insight has proven successful in several image restoration tasks \cite{Chen2015,Klatzer2016,Wang2016} in the last years. Defining a model architecture in this framework is typically done in the following steps: Given a corruption model and some data
(1) Choose a prior model; 
(2) Choose an inference model (possibly approximate);
(3) Unroll the inference procedure in time and interpret it as an RNN;
(4) Train the model parameters with back-propagation through time.\\
It becomes clear that in this framework the RNN architecture is fully determined by the choice of prior model and inference model. This imposes two major difficulties for good reconstruction models. First, it burdens practitioners with the choice of appropriate prior and inference models for the given data and task at hand, which typically requires a lot of expert knowledge from practitioners. Second, for complex data and inference tasks,  there are often no \textit{correct} choices of prior and inference. Instead there will typically be a trade-off between prior and inference procedure. This limitation is also present in the current RNN framework.

The goal of our work is to simplify the current RNN framework for learned iterative inference on the one hand, and to reduce its limitations on the other. We define a general class of models which we call ``Recurrent Inference Machines'' (RIM) that is able to learn an iterative inference algorithm without the need to explicitly specify a prior or a particular inference procedure, because they will be implicit in the model parameters. An RIM is an iterative map which includes the current reconstruction, a hidden memory state, and the gradient of the likelihood term which encodes information about the known generative process and measures how well we currently reproduce the measurements.

Training an RIM from a practitioners perspective then boils down to choosing an RNN architecture. Because RNNs are Turing complete \cite{Siegelmann1991,Siegelmann1995}, RIMs are a generalization of the models in \citet{Chen2015,Gregor2010,Wang2016,Zheng2015}.
RIMs are also related to a recent paper by \citet{Andrychowicz2016} that aims to train RNNs as optimizers for non-convex optimization problems. Though introduced with a different intention, RIMs can also be seen as a generalization of this approach, in which the model - in addition to the gradient information - is aware of the absolute position of a prediction in variable space(see equation \eqref{eq:update_equation}).

In this work we show that the change in viewpoint made possible by the RIM framework allows us to - with little domain knowledge - train models which outperform state-of-the-art models that have been hand engineered to perform specific tasks. We further show that RIMs perform much better than competing algorithms in across-task generalization, i.e. we can train an RIM on a deconvolution task and then apply it to an inpainting task by simply swapping the likelihood gradients but keeping the rest of the RIM parameters fixed. In the experiments we clearly demonstrate this across-task generalization ability. 

\section{Recurrent Inference Machines}\label{sec:RIM}
\begin{figure}[t]
\centering
\includegraphics[width=\textwidth]{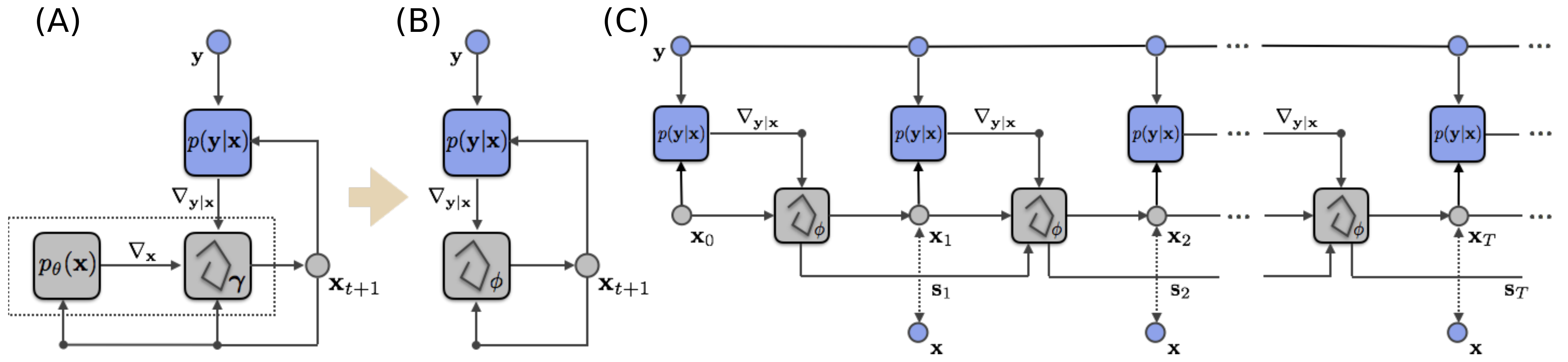}
\caption{
\textbf{(A)} \textit{Graphical illustration of the recurrent structure of MAP estimation} (compare equation \eqref{eq:basic_gradient}). The three boxes represent likelihood model $p(\y|\x)$, prior $p_\theta(\x)$, and update function $\Gamma$, respectively. 
In each iteration, likelihood and prior collect the current estimate of $\x$, to send a gradient to update function $\Gamma$ (see text). $\Gamma$ then produces a new estimate of $\x$. Grey boxes represent internal, data independent modules, while blue boxes represent external, data dependent modules.
\textbf{(B)} \textit{Model simplification.} Components $p_\theta(\x)$ and $\Gamma$ are merged into one model with trainable parameters $\phi$. The model iteratively produces new estimates through feedback from likelihood model $p(\y|\x)$ and previous updates. 
\textbf{(C)} \textit{A Recurrent Inference Machine unrolled in time.} Here we have added a hidden memory state $\s$. 
During training, estimates at each time step are subject to an error signal from the ground truth signal $\x$ (dashed two-sided arrows) in order to perform backpropagation. 
At test time, there is no error signal from $\x$.
}
\label{fig:model}
\end{figure}
In most inverse problems a closed-form map from observations $\y$ to signal $\x$ is intractable \cite{Nickisch2009,Figueiredo2007}. Instead, inverse problems are typically solved through use of iterative algorithms. Recurrent neural networks (RNN) can implement any algorithm because they are Turing complete \cite{Siegelmann1991,Siegelmann1995}. With the definition of Recurrent Inference Machines (RIM) we aim to establish a framework which allows us to apply RNNs for any kind of inverse problem. In the following, we will motivate the RIM framework from gradient-based inference methods.
%
Recall from equation \eqref{eq:map_optimization} that inverse problems can be interpreted in terms of probability such that optimization is an iterative approach to MAP inference. In its most simple form each consecutive estimate of $\x$ is then computed through a recursive function of the form
\begin{align}
\label{eq:basic_gradient}
\x_{t+1} &= \x_t + \gamma_t \nabla \Big(\log p\left(\y|\x \right) + \log p_{\theta}\left( \x \right) \Big) \left(\x_t\right)
\end{align}
where we make use of the fact that $p(\x| \y) \propto p(\y|\x)p_{\theta}(\x)$ and $\gamma_t$ is the step size or learning rate at iteration $t$. Further, $p(\y|\x)$ is the likelihood function for a given inference problem, and $p_{\theta}\left( \x \right)$ is a prior over signal $\x$. In many cases where either the likelihood term or the prior term deviate from standard models, optimization will not be convex. In contrast, the approach presented in this work is completely freed from ideas about convexity, as will be shown below.
The central insight of this work is that for a known forward model (likelihood model) update equation \eqref{eq:basic_gradient} can be generalized such that
\begin{align}
\label{eq:update_equation}
\x_{t+1} = \x_t + g_{\phi}(\nabla_{\y|\x},\x_{t})
 \end{align}
where we denote $\nabla \log p(\y|\x)(\x_{t})$ by $\nabla_{\y|\x}$ for readability and $\phi$ is a set of learnable parameters that govern the updates of $\x$. In this representation, prior parameters $\theta$ and learning rate parameters $\boldsymbol{\gamma}$ have been merged into one set of trainable parameters $\phi$. We can recover the original update equation \eqref{eq:basic_gradient} with
\begin{align}
 \label{eq:g_def}
g_{\phi}(\nabla_{\y|\x},\x_{t}) = \gamma_t  \left( \nabla_{\y|\x} + \nabla_{\x}\right)
 \end{align}
where we make use of $\nabla_{\x}$ to denote $\nabla \log p_\theta(\x)(\x_{t})$.
It will be useful to dissect the terms on the right-hand side of \eqref{eq:g_def} to make sense of the usefulness of the modification. 
First notice, that in equation \eqref{eq:basic_gradient} we never explicitly evaluate the prior, but only evaluate its gradient in order to perform updates. If never used, learning a prior appears to be unnecessary, and instead it appears more reasonable to directly learn a gradient function $\nabla_{\x} = f_{\theta}(\x_{t})  \in \mathbb{R}^d$. The advantage of working solely with gradients is that they do not require the evaluation of an (often) intractable normalization constant of $p_{\theta}(\x)$. A second observation is that the step sizes $\gamma_t$ are usually subject to either a chosen schedule or chosen through a deterministic algorithm such as a line search. That means the step sizes are always chosen according to a predefined model $\Gamma$. In order to make inference faster and improve performance we suggest to learn the model $\Gamma$ as well.\\
In \eqref{eq:update_equation} we have made prior $p_{\theta}(\x)$ and step size model $\Gamma$ implicit in function $g_{\phi}(\nabla_{\y|\x},\x_{t})$. We explicitly keep $\nabla_{\y|\x}$ as an input to \eqref{eq:update_equation} because - as opposed to $\Gamma$ and $p_{\theta}(\x)$ - it represents extrinsic information that is injected into the model. It allows for changes in the likelihood model $p(\y|\x)$ without the need to retrain parameters $\phi$ of the inference model $g_\phi$. \figref{fig:model} gives a visual summary of the insights from this section.

In many problem domains the range of values for variable $\x$ is naturally constrained. For example, images typically have pixels with strictly positive values. In order to model this constraint we make use of nonlinear link functions as they are typically used in neural networks, such that
$\x = \Psi(\z)$, where $\Psi(\cdot)$ is any differentiable link function and $\z$ is the space in which RIMs iterate. As a result $\x$ can be constrained to a certain range of values through $\Psi(\cdot)$, whereas iterations are performed in the unconstrained space of $\z$

We generalize \eqref{eq:update_equation} to adhere to the RNN framework by adding a latent memory variable $\s_t$. The central update equation of the RIM takes the form 
\begin{align}
\label{eq:update_equation_state}
\z_{t+1} &= \z_t + h_{\phi}\left(\nabla_{\y|\z},\z_{t},\s_{t+1}\right) \\
\s_{t+1} &= h^*_{\phi}\left(\nabla_{\y|\z},\z_{t},\s_{t}\right)
\end{align}
where $h^*_{\phi}(\cdot)$ is the update model for state variable $\s$. Intuitively, variable $\s$ will allow the procedure to have memory in order to track progression, curvature, approximate a preconditioning matrix $\mathbf{T}_{t}$ (such as in BFGS) and determine a stopping criterion among other things. The concept of a temporal memory is quite limited in classical inference methods, which will allow RIMs to have a potential advantage over these methods.

In order to learn a step-wise inference procedure it will be necessary to simulate the inference steps during training. I.e. during training, an RIM will perform a number of inference steps $T$. At each step the model will produce a prediction as depicted in figure \figref{fig:model}. Each of those predictions is then subject to a loss, which encourages the model 
to produce predictions that improve over time. In its simplest form, we can define a loss which is simply a weighted sum of the individual prediction losses at each time step such that
\begin{align}
	\mathcal{L}^{total}(\phi) = \sum_{t=1}^T w_t \mathcal{L}(\x_t(\phi),\x)
\end{align}
is the total loss. Here, $\mathcal{L}(\cdot)$ is a base loss function such as the mean square error, $w_t$ is a positive scalar and $\x_t(\phi)$ is a prediction at time $t$. In this work, we follow \citet{Andrychowicz2016} in setting $w_t = 1$ for all time steps.
%
%
%
%
%


\section{Experimental Results}
We evaluate our method on various kinds of image restoration tasks which can each be formulated in terms of a linear measurement problem (see \eqref{eq:linear_mesaurement}).
We first analyze the properties of our proposed method on a set of restoration tasks from random projections. Later we compare our model on two well known image restoration tasks: image denoising and image super-resolution.

\paragraph{Models}
If not specified otherwise we use the same RIM architecture for all experiments presented in this work. The chosen RIM consists of three convolutional hidden layers and a final convolutional output layer. All convolutional filters were chosen to be of size 3 x 3 pixels. The first hidden layer consists of convolutions with stride 2 (64 features) and a tanh nonlinearity. The second hidden layer represents the RNN part of the model. We chose a gated recurrent unit (GRU) \citep{Chung2014} with 256 features. The third hidden layer is a transpose convolution layer with 64 features which aims to recover the original image dimensions of the signal, followed again a tanh nonlinearity. All models were trained on a fixed number of iterations of 20 steps. All methods were implemented in Tensorflow\footnote{https://www.tensorflow.org}.

\paragraph{Data}
All experiments were run on the BSD-300 data set \citep{MartinFTM01}\footnote{https://www2.eecs.berkeley.edu/Research/Projects/CS/vision/bsds/}. For training we extracted patches of size 32 x 32 pixels with stride 4 from the 200 training images available in the data set. In total this amounts to a data set of about 400 thousand image patches with highly redundant information. All models were trained over only two epochs. Validation was performed on a held-out data set of 1000 image patches.
For testing we either used the whole test set of 100 images from BSDS-300 or we used only a subset of 68 images which was introduced by \citet{Roth2005} and which is commonly used in the image restoration community \footnote{http://www.visinf.tu-darmstadt.de/vi\_research/code/foe.en.jsp}.

\begin{figure}[t]
\includegraphics[width=\textwidth]{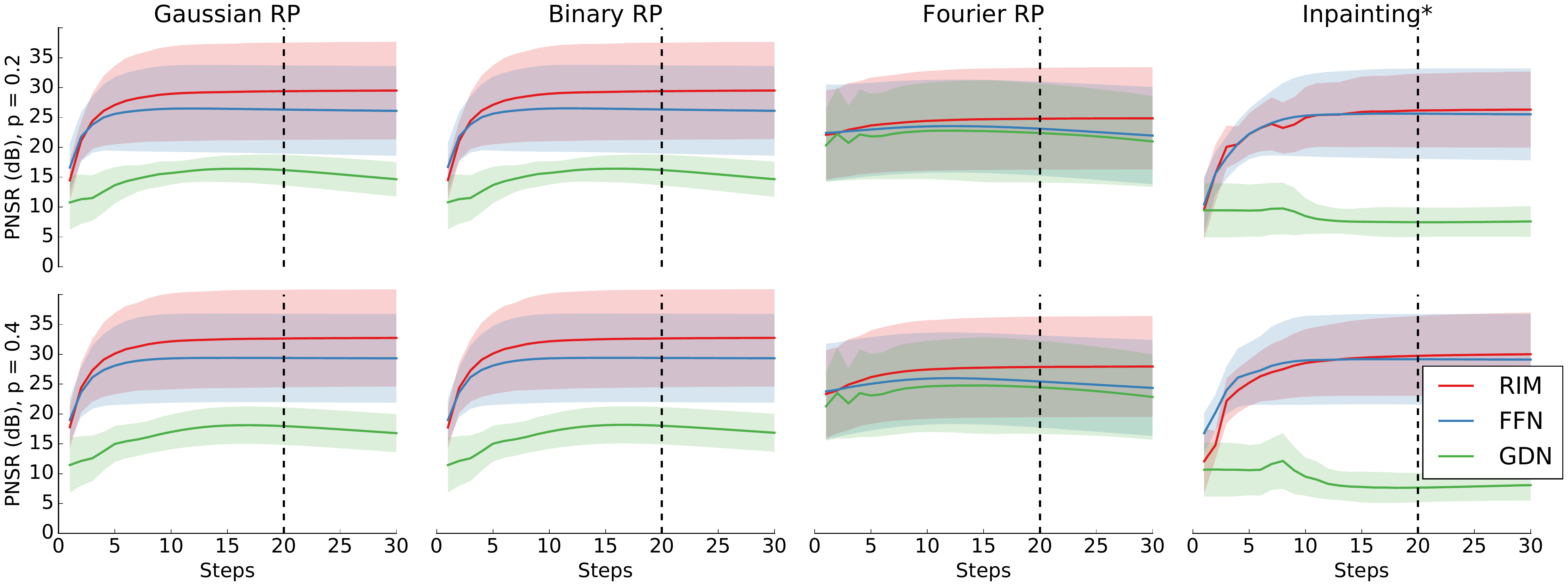}
\caption{\textit{Reconstruction performance over time on random projections.} Shown are results of the three reconstruction tasks from random projections (see text) on 5000 random patches from the BSD-300 test set. Value of p represent the reduction in dimensionality through the random projection. Noise standard deviation was chosen to be $\sigma=1$. Solid lines correspond to the mean peak signal-to-noise-ratio (PSNR) over time, and shaded areas correspond to one standard deviation around the mean. Vertical dashed lines mark the last time step that was used during training.}
\label{fig:random_projections}
\end{figure}

\paragraph{Image Restoration}
All tasks addressed in this work assume a linear measurement problem
of the form as described in equation \eqref{eq:linear_mesaurement} with additive (isotropic) Gaussian noise. In this case, we write the likelihood-gradient as
\vspace{-1em}
\begin{align}
\nabla_{\y|\x} = \frac{1}{\sigma^2 + \epsilon} \A^T (\y - \A \x)
\end{align}
where $\sigma^2$ is the noise variance. For very small $\sigma$ the gradient diverges. To make the gradient more stable also for small $\sigma$ we add $\epsilon = \text{softplus}(\phi_\epsilon)$, where $\phi_\epsilon$ is a trainable parameter.
As a link function $\Psi$ we chose the logistic sigmoid nonlinearity\footnote{All training data was rescaled to be in the range $[0,1]$} and we used the mean square error as training loss.

\subsection{Multi-task learning with Random Projections} \label{sec:multitask}
To test convergence properties and the model components of the RIM, we first trained the model to reconstruct gray-scale image patches from noisy random projections. We consider three types of random projection matrices: (1) Gaussian ensembles with elements drawn from a standard normal distribution, (2) binary ensembles with entries of values $\{-1, 1\}$ drawn from a Bernoulli distribution with $p=0.5$, and (3) Fourier ensembles with randomly sampled rows from a Fourier matrix (deconvolution task)\cite{Donoho2006a}.


We trained three models on these tasks: (1) a Recurrent Inference Machine (RIM) as described in \ref{sec:RIM}, (2) a gradient-descent network (GDN) which does not use the current estimate as an input (compare \citet{Andrychowicz2016}), and (3) a feed-forward network (FFN) which uses the same inputs as the RIM but where we replaced the GRU unit with a ReLu layer in order to remove hidden state dependence. Model (2) and (3) are simplifications of RIM in order to test the influence of each of the removed model components on prediction performance. Each model was trained to perform all three reconstruction tasks under the same set of learned parameters.

Figure \ref{fig:random_projections} shows the reconstruction performance of all three models on random projections. In all tasks the RIM clearly outperforms both other models, showing overall consistent convergence behavior. The FFN performs well on easier tasks but starts to show degrading performance over time on more difficult tasks. This suggests that the state information of RIM plays an important role in the convergence behavior as well as overall performance. The GDN shows the worst performance among all three models. For all tasks, the performance of GDN starts to degrade clearly after the 20 time steps that were used during training. We hypothesize that the model is able to compensate some of the missing information about the current estimate of $\x$ through state variable $\s$ during training, but the model is not able to transfer this ability to episodes with more iterations.
\begin{figure}[t]
\centering
\begin{subfigure}{\textwidth}
\includegraphics[width=\textwidth]{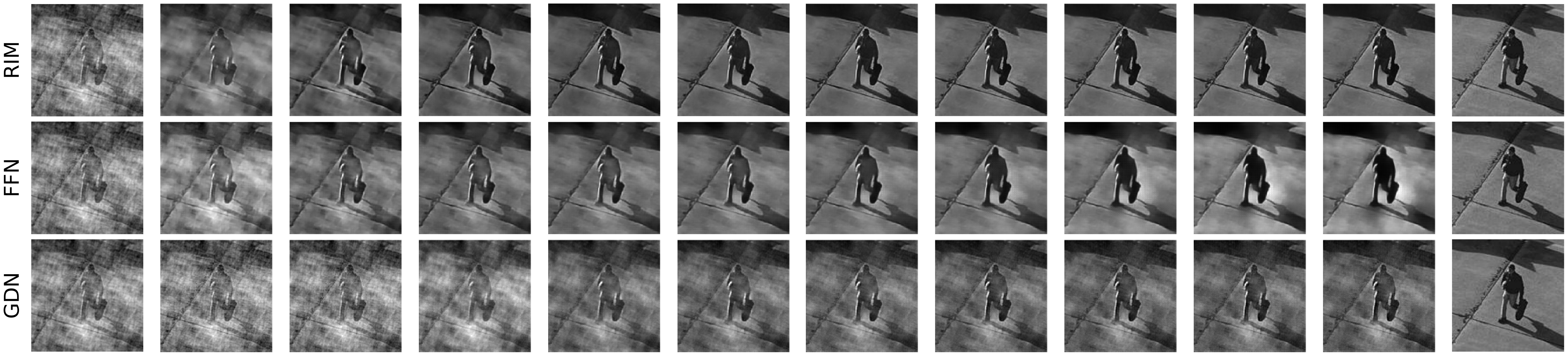}
\caption{Fourier ensemble with $p = 0.4$.}
\label{fig:reconst_fourier}
\end{subfigure}
\begin{subfigure}{\textwidth}
\includegraphics[width=\textwidth]{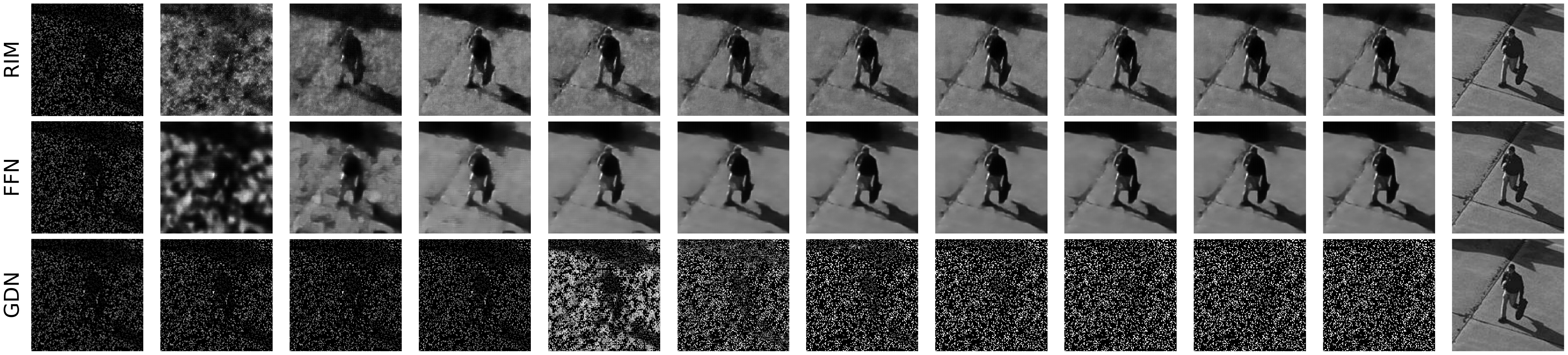}
\caption{Inpainting task with $p = 0.2$.}
\label{fig:reconst_inpaint}
\end{subfigure}
\captionsetup{width=\textwidth}
\caption{\textit{Image restoration of a 128 x 128 pixel image.} Models as described in \ref{sec:multitask}. Leftmost image is the starting guess $\A^T \y$, rightmost image is the ground truth image, and in between are every two steps of the model outputs. p is the fraction of random projection.}
\label{fig:iterative_reconst}
\end{figure}

We further evaluated all three models on an out-of-training task: image inpainting. In an image inpainting task pixels are randomly removed from an image. Both the RIM and FFN are able to transfer knowledge to the new task, whereas the GDN fails to generalize. This further emphasizes the importance of using the current estimate of $\x$ as an input to the model, suggesting that it enables these models to learn a prior.
Figure \ref{fig:iterative_reconst} shows some exemplary results of the learned iterative algorithms. Both, RIM and FFN are able to recover the input image over time, while the RIM consistently displays the best results. The figure also emphasizes generalization to larger images: models were trained on 32 x 32 pixel image patches, but are performing reconstruction on a 128 x 128 image patch.

These results suggests that both the current estimate as well as the recurrent state carry useful information for performing inference. They further show that the RIM framework allows us to transfer knowledge to new tasks without the need for retraining. In the following, we will therefor only consider fully fledged RIMs.
\subsection{Image Denoising}
\begin{wraptable}{r}{4.3cm}
\centering
\caption{\textit{Color denoising.} Denoising of 68 BSD-300 test images, $\sigma = 25$ after 8-bit quantization. Results for RTF-5 and CBM3D  adopted from \citet{Schmidt2016}. Results for full 100 test images in parenthesis.}
\begin{tabular}{@{}lc@{}}
\toprule
Method & PSNR          \\ \midrule
CBM3D \cite{Dabov2007b} & $30.18$        \\
RTF-5 \cite{Schmidt2016} & $30.57$        \\
\textbf{RIM (ours)}    & $\mathbf{30.84} (30.67)$ \\ \bottomrule
\end{tabular}
\label{tab:denoise_color}
\end{wraptable}
After evaluating our model on 32 x 32 pixel image patches we wanted to see how reconstruction performance generalizes to full sized images and to another out of domain problem. We chose to reuse the RIM that was trained on the random projections task to perform image denoising (further called RIM-3task). To test the hypothesis that inference should be trained task specific, we further trained a model RIM-denoise solely on the denoising task. Table \ref{tab:denoise_gray} shows the denoising performance on the BSD-300 test set for both models as compared to state-of-the-art methods in image denoising. The RIM-3task model shows very competitive results with other methods on all noise levels. This further exemplifies that the model indeed has learned something reminiscent of a prior, as it was never directly trained on this task. The RIM-denoise model further improves upon the performance of RIM-3task and it outperforms most other methods on all noise levels. This is to say that the same RIM was used to perform denoising on different noise levels, and this model does not require any hand tuning after training. 
\begin{table}[t]
\centering
\caption{\textit{Gray-scale denoising.} Shown are mean PSNR for 68 test images from \citet{Roth2005}, numbers in parenthesis correspond to performance on all 100 test images from BSD-300. 68 image performance for BM3D, LSSC, EPLL, and opt-MRF adopted from \citet{Chen2013}. Performances on 100 images adopted from \citet{Burger2013}. 68 image performance on MLP \citep{Burger2012}, RTF-5 \citep{Schmidt2016} and all quantized results adopted from \citet{Schmidt2016}.
}
\setlength{\aboverulesep}{0pt}
\setlength{\belowrulesep}{0pt}
\setlength{\extrarowheight}{.7ex}
\resizebox{\textwidth}{!}{%
\begin{tabular}{@{}lcccccc@{}}
\toprule
                          & \multicolumn{3}{c}{Not Quantized}  & \multicolumn{3}{c}{8-bit Quantized}           \\
                          \cmidrule(r){2-4} \cmidrule(r){5-7}
\multicolumn{1}{c}{$\sigma$} & 15            & 25            & 50            & 15            & 25            & 50            \\
\cmidrule(r){1-1} \cmidrule(r){2-4} \cmidrule(r){5-7}
BM3D   \cite{Dabov2007}                   & $31.08$         & $28.56 (28.35)$ & $25.62 (25.45)$ &               & $28.31$         &               \\
LSSC \cite{Mairal2009}                      & $31.27$         & $28.70$         & $25.72$         &               & $28.23$         &               \\
EPLL  \cite{Zoran2011}                    & $31.19$         & $28.68 (28.47)$ & $25.67 (25.50)$ &               &               &               \\
opt-MRF \cite{Chen2013}                  & $31.18$         & $28.66$         & $25.70$         &               &               &               \\
MLP \cite{Burger2012}                       &               & $28.85 (28.75)$ & $(25.83)$       &               &               &               \\
RTF-5 \cite{Schmidt2016}                    &               & $28.75$         &               &               & $28.74$         &               \\ 
\hdashline
\textbf{RIM-3task}                 & $31.19 (30.98)$ & $28.67 (28.45)$ & $25.78 (25.59)$ & $31.06 (30.88)$ & $28.41 (28.24)$ & $24.86 (24.73)$ \\
\textbf{RIM-denoise}               & $\mathbf{31.31} (31.10)$ & $\mathbf{28.91} (28.72)$ & $\mathbf{26.06} (25.88)$ & $\mathbf{31.25} (31.05)$ & $\mathbf{28.76} (28.58)$ & $\mathbf{25.27} (25.14)$ \\ \bottomrule
\end{tabular}}
\label{tab:denoise_gray}
\end{table}
\begin{figure}[t]
    \centering
    \begin{subfigure}[b]{0.35\textwidth}
	    \caption{Ground truth}
        \includegraphics[width=\textwidth]{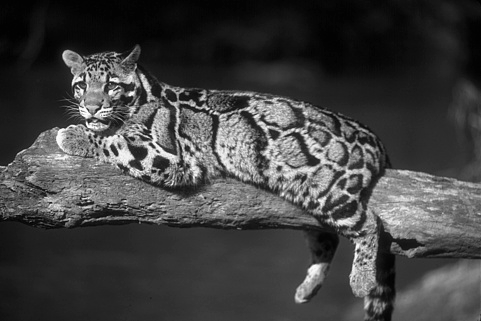}
    \end{subfigure}
    \begin{subfigure}[b]{0.35\textwidth}
    	\caption{Noisy image, 14.88dB}
        \includegraphics[width=\textwidth]{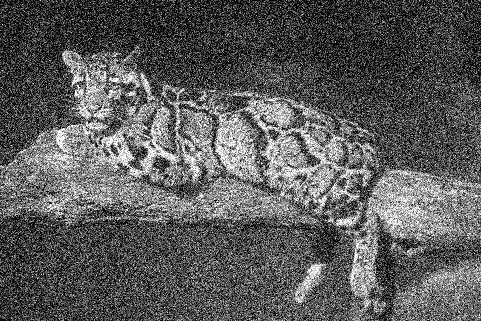}
    \end{subfigure}
    \\
    \vspace*{0.05cm}
    \begin{subfigure}[b]{0.35\textwidth}
        \includegraphics[width=\textwidth]{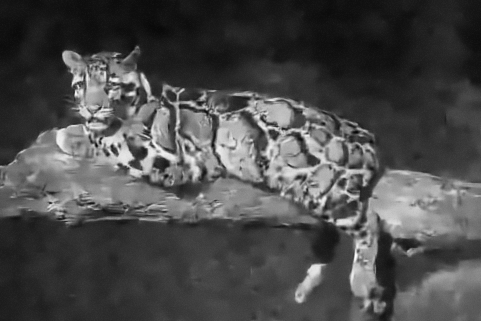}
        \caption{EPLL, 25.68dB}
    \end{subfigure}
    \begin{subfigure}[b]{0.35\textwidth}
	    \includegraphics[width=\textwidth]{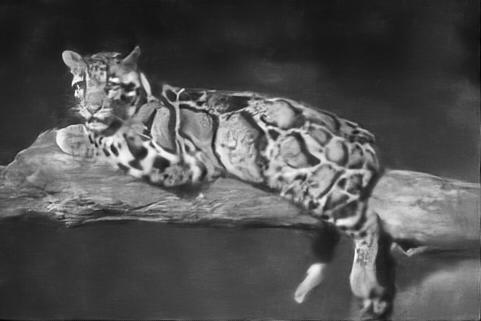}
        \caption{RIM, 25.91dB}
    \end{subfigure}
    \captionsetup{width=0.75\textwidth}
    \caption{\textit{Denoising performance on example image used in \citet{Zoran2011}.} $\sigma = 50$. Noisy image was 8-bit quantized before reconstruction.}
    \label{fig:denoising_gray}
\end{figure}
Table \ref{tab:denoise_gray} shows denoising performance on images that have been 8-bit quantized after adding noise (see \citet{Schmidt2016}). In this case performance slightly deteriorates for both models, though still being competitive with state-of-the-art methods. This effect could possibly be accommodated through further training, or by adjusting the forward model. Figure \ref{fig:denoising_gray} gives some qualitative results on the denoising performance for one of the test images from BSD-300 as compared to the method from \citet{Zoran2011}. RIM is able to produce more naturalistic images with less visible artifacts. The state variable in our RIM model allows for a growing receptive field size over time, which could explain the good long range interactions that the model exhibits.

Many denoising algorithms are solely tested on gray-scale images. Sometimes this is due to additional difficulties that multi-channel problems bring for some inference approaches. To show that it is straightforward to apply RIMs to multi-channel problems we trained a model to denoise RGB images. The denoising performance can be seen in table \ref{tab:denoise_color}. The model is able to exploit correlations across color channels which allows for an additional boost in reconstruction performance.
\subsection{Image Super-resolution}
\begin{figure}[t]
    \centering
    \begin{subfigure}[b]{0.32\textwidth}
	    \caption{Original Image}
\adjincludegraphics[clip,trim={.1\width} {.1\height} {.1\width} {.2\height},width=\textwidth]{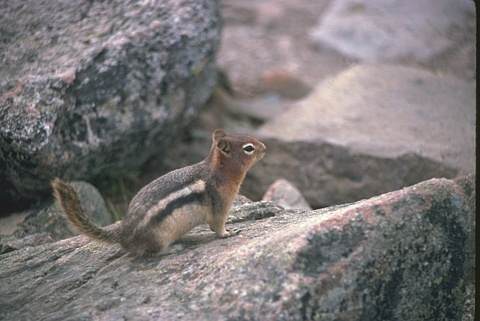}\llap{\adjustbox{trim={.4\width} {.4\height} {0.4\width} {.4\height},clip}{\includegraphics[width=2\textwidth]{results/img_013_SRF_3_HR.png}}}
    \end{subfigure}
    \begin{subfigure}[b]{0.32\textwidth}
	    \caption{Bicubic: $30.43/0.8326$}
  \adjincludegraphics[clip,trim={.1\width} {.1\height} {.1\width} {.2\height},width=\textwidth]{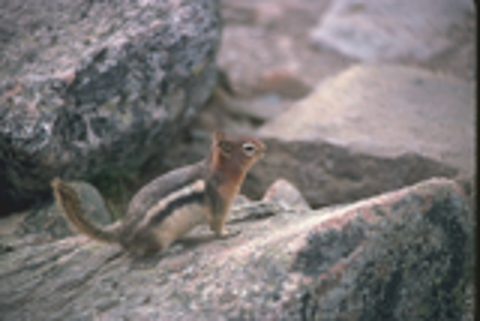}\llap{\adjustbox{trim={.4\width} {.4\height} {0.4\width} {.4\height},clip}{\includegraphics[width=2\textwidth]{results/img_013_SRF_3_bicubic.png}}}
    \end{subfigure}
    \begin{subfigure}[b]{0.32\textwidth}
	    \caption{SRCNN: $31.34 / 0.8660$}
  \adjincludegraphics[clip,trim={.1\width} {.1\height} {.1\width} {.2\height},width=\textwidth]{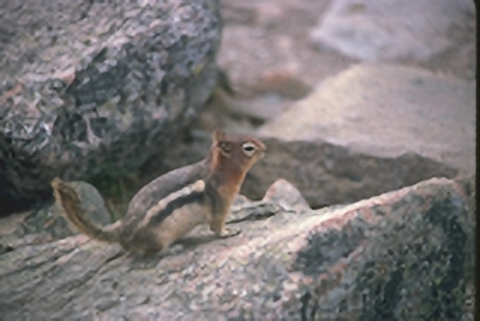}\llap{\adjustbox{trim={.4\width} {.4\height} {0.4\width} {.4\height},clip}{\includegraphics[width=2\textwidth]{results/img_013_SRF_3_SRCNN.png}}}
    \end{subfigure}
    \\
    \vspace*{0.05cm}
    \begin{subfigure}[b]{0.32\textwidth}
	    \adjincludegraphics[clip,trim={.1\width} {.1\height} {.1\width} {.2\height},width=\textwidth]{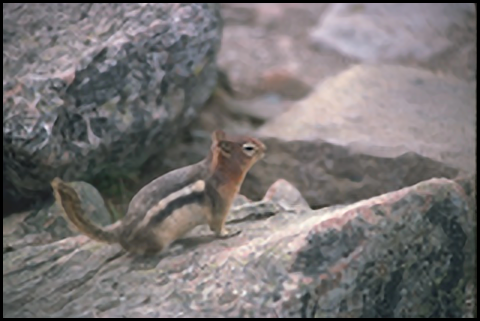}\llap{\adjustbox{trim={.4\width} {.4\height} {0.4\width} {.4\height},clip}{\includegraphics[width=2\textwidth]{results/img_013_SRF_3_A+.png}}}
        \caption{A+: $31.43/0.8676$}
    \end{subfigure}
    \begin{subfigure}[b]{0.32\textwidth}
  \adjincludegraphics[clip,trim={.1\width} {.1\height} {.1\width} {.2\height},width=\textwidth]{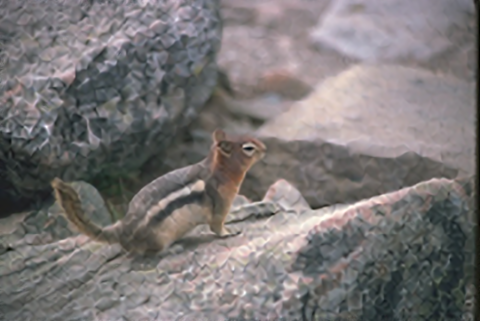}\llap{\adjustbox{trim={.4\width} {.4\height} {0.4\width} {.4\height},clip}{\includegraphics[width=2\textwidth]{results/img_013_SRF_3_SelfExSR.png}}}
  \caption{SelfExSR: $31.18/0.8656$}
    \end{subfigure}
    \begin{subfigure}[b]{0.32\textwidth}
  \adjincludegraphics[clip,trim={.1\width} {.1\height} {.1\width} {.2\height},width=\textwidth]{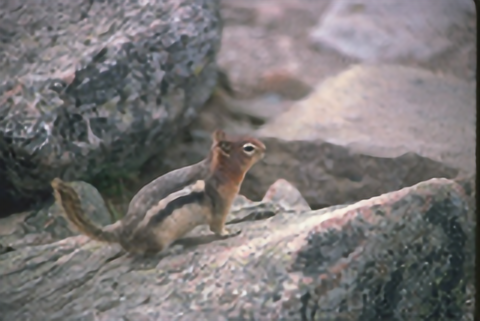}\llap{\adjustbox{trim={.4\width} {.4\height} {0.4\width} {.4\height},clip}{\includegraphics[width=2\textwidth]{results/img_013_SRF_3_RIM.png}}}
  \caption{RIM: $\mathbf{31.59}/\mathbf{0.8712}$}
\end{subfigure}\caption{\textit{Super-resolution example with factor 3.} Comparison with the same methods as in table \ref{tab:superres}. Reported numbers are PSNR/SSIM. Best results in bold.}
    \label{fig:superres}
\end{figure}

We further tested our approach on the well known image super-resolution task. We trained a single RIM \footnote{The architecture of this model was slightly simplified in comparison to the previous problems. Instead of strided convolutions, we chose \textit{a trous} convolutions. This model is more flexible and used only about $500.000$ parameters.} on 36 x 36 pixel image patches from the BSD-300 training set to perform image super-resolution for factors 2, 3, and 4\footnote{We reimplemented MATLABs bicubic interpolation kernel in order to apply a forward model (sub-sampling) in TensorFlow which agrees with the forward model in \citet{Huang2015}.}. We followed the same testing protocol as in \citet{Huang2015}, and we used the test images that were retrieved from their website \footnote{https://sites.google.com/site/jbhuang0604/publications/struct\_sr}. Table \ref{tab:superres} shows a comparison with some state-of-the-art methods on super-resolution for the BSD-300 test set. \figref{fig:superres} shows a qualitative example of super-resolution performance. The other deep learning method in this comparison, SRCNN \cite{Dong2014}, is outperformed by RIM on all scales. Interestingly SRCNN was trained for each scale independently whereas we only trained one RIM for all scales. The chosen RIM has only about $500.000$ parameters which amounts to about $2$MB of disk space, which makes this architecture very attractive for mobile computing.
\begin{table}[t]
\centering
\caption{\textit{Color image super-resolution.} Mean and standard deviation (of the mean) of Peak Signal-to-Noise Ratio (PSNR) and Structural Similarity Index (SSIM) \cite{Wang2004}. Standard deviation of the mean was estimated from $10.000$ bootstrap samples. Test protocol and images taken from \citet{Huang2015}. Only the three best performing methods from \citet{Huang2015} were chosen for comparison. Best mean values in bold.}
\setlength{\aboverulesep}{0pt}
\setlength{\belowrulesep}{0pt}
\setlength{\extrarowheight}{1ex}
\resizebox{\textwidth}{!}{%
\begin{tabular}{@{}llccccc}
\toprule
Metric                &Scale &Bicubic             &SRCNN \cite{Dong2014}              & A+ \cite{Timofte2015}                 & SelfExSR \cite{Huang2015}           &\textbf{RIM (Ours)}                 \\ \cmidrule(r){1-1} \cmidrule(r){2-2} \cmidrule(r){3-7}
\multirow{3}{*}{PSNR} & 2x    & $29.55 \pm 0.35$    & $31.11 \pm 0.39$    & $31.22 \pm 0.40$    & $31.18 \pm 0.39$    & $\mathbf{31.39} \pm 0.39$    \\
                      & 3x    & $27.20 \pm 0.33$    & $28.20 \pm 0.36$    & $28.30 \pm 0.37$    & $28.30 \pm 0.37$    & $\mathbf{28.51} \pm 0.37$    \\
                      & 4x    & $25.96 \pm0.33$     & $26.70 \pm 0.34$    & $26.82 \pm 0.35$    & $26.85 \pm 0.36$    & $\mathbf{27.01} \pm 0.35$    \\ \cmidrule(r){1-1} \cmidrule(r){2-2} \cmidrule(r){3-7}
\multirow{3}{*}{SSIM} & 2x    & $0.8425 \pm 0.0078$ & $0.8835 \pm 0.0062$ & $0.8862 \pm 0.0063$ & $0.8855 \pm 0.0064$ & $\mathbf{0.8885} \pm 0.0062$ \\
                      & 3x    & $0.7382 \pm 0.0114$ & $0.7794 \pm 0.0102$ & $0.7836 \pm 0.0104$ & $0.7843 \pm 0.0104$ & $\mathbf{0.7888} \pm 0.0101$ \\
                      & 4x    & $0.6672 \pm 0.0131$ & $0.7018 \pm 0.0125$ & $0.7089 \pm 0.0125$ & $0.7108 \pm 0.0124$ & $\mathbf{0.7156} \pm 0.0125$ \\ \bottomrule
\end{tabular}%
}
\label{tab:superres}
\end{table}

\section{Discussion}
In this work, we introduce a general learning framework for solving inverse problems with deep learning approaches. We establish this framework by abandoning the traditional separation between model and inference. Instead, we propose to learn both components jointly without the need to define their explicit functional form. This paradigm shift enables us to bridge the gap between the fields of deep learning and inverse problems. A crucial and unique quality of RIMs are their ability to generalize across tasks without the need to retrain. We convincingly demonstrate this feature in our experiments as well as state of the art results on image denoising and super-resolution. 

We believe that this framework can have a major impact on many inverse problems, for example in medical imaging and radio astronomy. Although we have focused on linear image reconstruction tasks in this work, the framework can be applied to inverse problems of all kinds, such as non-linear inverse problems.
\newpage

\newpage
\small
\bibliographystyle{unsrtnat} 
\bibliography{references}

\end{document}